\documentclass{article}

\PassOptionsToPackage{numbers,compress}{natbib}
\usepackage[preprint]{neurips_2026}

\usepackage[utf8]{inputenc}
\usepackage[T1]{fontenc}
\usepackage[hyphens]{url}
\usepackage{hyperref}
\hypersetup{hidelinks}
\usepackage{float}
\usepackage{caption}
\usepackage{algorithm}
\usepackage{algorithmic}
\usepackage{newfloat}
\usepackage{listings}
\DeclareCaptionStyle{ruled}{labelfont=normalfont,labelsep=colon,strut=off}
\floatstyle{ruled}
\newfloat{listing}{tb}{lst}{}
\floatname{listing}{Listing}

\usepackage{booktabs}
\usepackage{enumitem}
\usepackage{amsmath}
\usepackage{mathtools}
\DeclareMathOperator*{\argmax}{arg\,max}

\usepackage{graphicx}
\usepackage{wrapfig}
\usepackage{pgfplots}
\usepgfplotslibrary{groupplots}
\usetikzlibrary{calc}
\pgfplotsset{compat=1.18}

\usepackage{multirow}
\usepackage{multicol}
\usepackage{makecell}
\usepackage{bm}
\usepackage{xspace}
\usepackage{xcolor}
\usepackage[most]{tcolorbox}
\usepackage{titlesec}
\titlespacing*{\subsection}{0pt}{1.1ex plus .2ex minus .1ex}{0.55ex plus .1ex}
\titlespacing*{\subsubsection}{0pt}{0.95ex plus .2ex minus .1ex}{0.45ex plus .1ex}
\titlespacing*{\paragraph}{0pt}{0.55ex plus .1ex minus .1ex}{0.7em}

\newcommand{\ourmethod}{\textsc{CurateEvo}\xspace}
\definecolor{softred}{RGB}{250,100,100}
\definecolor{softgreen}{RGB}{56,118,29}
\definecolor{softblue}{RGB}{100,150,200}
\definecolor{softgray}{RGB}{150,150,150}

\newtcolorbox[auto counter, number within=section]{prompt}[2][]{%
  colback=white,
  colframe=softblue!150,
  width=\textwidth,
  arc=3mm,
  boxrule=0.8mm,
  title=\normalsize #2,
  breakable,
  fonttitle=\small,
  fontupper=\footnotesize,
  #1
}

\title{\ourmethod: Data-Curation Evolving for Agentic Post-Training}

\author{
    Dingzirui Wang, Xuanliang Zhang, Keyan Xu Qingfu Zhu, Wanxing Che\thanks{Corresponding Author} \\
    Harbin Institue of Technology \\
    \{dzrwang, xuanliangzhang, kyxu, qfzhu, car\}@ir.hit.edu.cn
}

\begin{document}
    \maketitle

    \begin{abstract}
        Large language model (LLM) agents require post-training methods that can improve long-horizon decision making from environment feedback. However, existing agentic post-training pipelines often treat data curation as a fixed preprocessing step, focusing mainly on data augmentation while neglecting filtering, refinement, and adaptation to downstream failures. We propose \ourmethod, a failure-driven dynamic evolution framework for agentic post-training data curation. \ourmethod represents the curation strategy as executable code and iteratively rewrites it using failed trajectories from a held-out development set. At each epoch, the evolved strategy transforms a fixed raw corpus into supervised fine-tuning data, reinforcement learning data, and an inference-time memory bank. The evolution process first improves effectiveness by diagnosing recurring failure modes and augmenting, filtering, or refining data accordingly, and then improves efficiency by pruning redundant or low-utility training turns under a cost-aware objective. Experiments on ACEBench-Agent, BFCL-V4, and $\tau^2$-Bench under both labeled and wild-data settings show that \ourmethod consistently outperforms prior curation methods, improving average scores by $3.2$ and $2.7$ points, respectively. Further analyses demonstrate that \ourmethod is compatible with different post-training recipes and substantially reduces curation overhead.
    \end{abstract}
    
    \section{Introduction}
        Although large language models (LLMs) have demonstrated strong language understanding and reasoning abilities, text generation alone is often insufficient for solving complex real-world tasks.
LLM agents extend LLMs into interactive systems by incorporating perception, planning, memory, and tool use, enabling models to interact with external environments and take concrete actions.
Modern agent systems typically use an LLM as the core decision-making module, which continuously selects actions according to the task goal and environment feedback.
However, agent tasks are often characterized by long-horizon interactions, sparse feedback, and high decision uncertainty, making it difficult for standard supervised learning to fully optimize the interaction process.
Agentic post-training, which directly improves decision policies using environment feedback, has therefore become an important approach for enhancing the agent ability of LLMs.

Prior studies have shown that data quality is a key factor that determines the performance of agentic post-training \cite{qi2025webrl,jin2025searchr}.
As a result, many recent methods curate training data according to model feedback, for example, by collecting failed trajectories from evaluation~\cite{song-etal-2024-trial,wang-etal-2025-nat} or by interacting with simulated environments~\cite{zala2024envgen,khan2025dataenvgym} to acquire data on which the current model performs poorly.
\pagebreak[4]
\begin{wrapfigure}{r}{0.43\textwidth}
    \vspace{-0.15em}
    \centering
    \includegraphics[width=\linewidth]{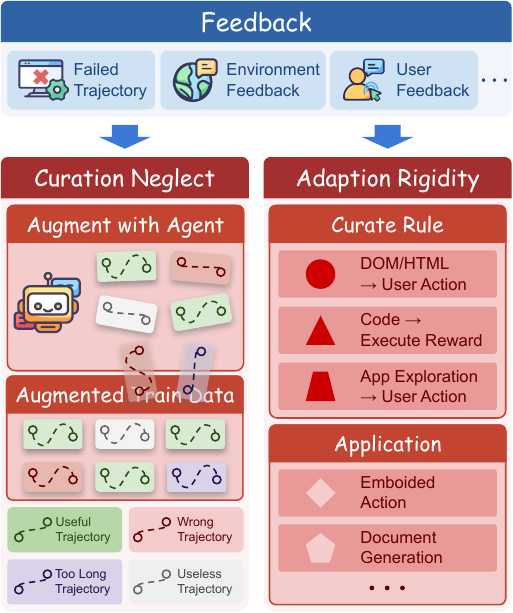}
    \caption{
        Two limitations of existing agentic post-training data curation methods.
        \textit{Curation Neglect} denotes overlooking the curation operation like filter and refine, mixing low quality data.
        \textit{Adaption Rigidity} denotes the curation process is fixed, which cannot be adapted to different applications flexibly.
    }
    \vspace{-1.2em}
    \label{fig:motivation}
\end{wrapfigure}
Despite their effectiveness, existing data curation methods for agentic post-training still suffer from two limitations, as shown in Figure~\ref{fig:motivation}.
\textit{(i) Curation Neglect}: Most methods mainly focus on augmenting additional data, while prior studies suggest that filtering~\cite{xia2024less} and refining~\cite{hu-etal-2025-webcot} data can further improve agentic post-training performance while reducing training costs.
\textit{(ii) Adaptation Rigidity}: Existing methods usually rely on fixed data processing pipelines, which are difficult to adapt to out-of-distribution downstream applications.

To address the above limitations, we propose \ourmethod, a dynamic evolution framework to handle various data-curation operations for agentic post-training.
Instead of treating data curation as a static process, \ourmethod improves the data curation strategy through multi-epoch evolution.
At each evolution epoch, \ourmethod uses failed trajectories on a held-out dev set as feedback, treats the data curation code as the optimization target, and aims to improve agent effectiveness while reducing the scale of training turns.
Specifically, \ourmethod first extracts failure modes from failed trajectories and revises the curation code to generate, filter, or refine data that targets these failure modes.
After addressing effectiveness-related failure modes, \ourmethod further improves efficiency by modifying the curation code to remove low-utility data, prune redundant data, and truncate overly long trajectories, thereby reducing the total number of interaction turns used for training.

Compared with prior work, \ourmethod has two main advantages.
\textit{(i) Curation Awareness}: \ourmethod jointly considers data augmentation, filtering, and refinement during curation, which allows it to improve agentic post-training performance while reducing training costs.
\textit{(ii) Adaptation Flexibility}: \ourmethod dynamically revises the data curation according to evaluation feedback, making it better suited for downstream out-of-distribution applications.

To evaluate the generality of \ourmethod, we conduct experiments under both labeled and wild training settings.
The labeled setting studies how to curate human-annotated training data, while the wild setting directly processes real user interaction trajectories as training data.
Experimental results show that \ourmethod improves over baseline methods by $3.2$ and $2.7$ in the labeled and wild settings, respectively, demonstrating its effectiveness.
Further analysis shows that \ourmethod can be combined with existing agentic post-training methods to achieve an additional improvement of $21.3$, suggesting that \ourmethod enhances agentic post-training by improving the quality and efficiency of training data.

Our contributions are summarized as follows:
\begin{itemize}[leftmargin=*,nosep]
    \item We propose \ourmethod, a dynamic evolution framework to handle various data-curation operations for agentic post-training to enhance the effectiveness and efficiency of agentic post-training.
    \item We evaluate \ourmethod under both labeled and wild training settings, where it improves over baseline methods by $3.2$ and $2.7$, respectively, demonstrating its effectiveness across different data sources.
    \item We provide further analysis showing that \ourmethod can be combined with existing agentic post-training methods for additional performance gains, highlighting its practicality in real-world agent application scenarios.
\end{itemize}

    \section{Related Work}
        Data curation for agentic post-training refers to the process of augmenting, filtering, and refining interaction data for supervised warm-up and reinforcement learning of language agents. 
It is particularly important because agent trajectories are often long-horizon, heterogeneous across environments and action spaces, noisy, and dominated by sparse successes and frequent failures. 
Consequently, the curated data distribution directly affects exploration efficiency, reward learning, and final agent performance.

Existing studies reveal a gradual shift from static trajectory collection to feedback-driven data evolution. 
Early efforts established the basic recipe by constructing agent tasks or collecting high-quality trajectories for web navigation, tool use, and general agent behaviors, providing important foundations but remaining limited in scalability~\cite{yao2022webshop,deng2023mindweb,chen2024fireact,zeng-etal-2024-agenttuning}. 
As agent data became larger and more diverse, later work focused on making heterogeneous trajectories more usable for training through automatic tool-use data construction, unified trajectory formats, modular data pipelines, corpus redesign, negative-sample design, large-scale trajectory banks, and rejection-sampling or curriculum-style refinement~\cite{qin2024toolllm,yin-etal-2024-agent,zhang2024agentohanadesignunifieddata,chen-etal-2024-agent,song-etal-2024-agentbank,lai-etal-2024-autowebglm}. 
More recent approaches further close the loop between data curation and policy optimization, where agents collect trajectories in executable environments, receive verifiable or environment-derived feedback, and use this feedback to guide task generation, rollout filtering, reward modeling, and multi-turn reinforcement learning~\cite{xi2026agentgymrl,trabucco2026insta,qi2025webrl,wei-etal-2025-webagent,wang2025ragenunderstandingselfevolutionllm}. 
A parallel and increasingly important direction is to exploit failed or partial trajectories rather than discard them, for example, by incorporating negative examples, selecting critical decision steps, mining useful actions from failed expert rollouts, adaptively curating high-value samples, or standardizing trajectories through unified data protocols~\cite{wang2024learning,chen-etal-2025-atlas,lan2026exploring,li2026efficient,song2026agent}. 
Overall, agentic post-training data curation is moving beyond manual selection and rule-based cleaning toward closed-loop, feedback-aware, and failure-conscious data evolution.

Despite these advances, existing agentic post-training data curation methods remain limited, as they only focus on augmentation without jointly optimizing data filtering and refinement under training-cost constraints. 
Moreover, fixed curation pipelines are hard to adapt to new tasks and failure modes. 
To overcome these limitations, \ourmethod proposes evolving data curation, which is iteratively improved using failed trajectories as feedback.
This enables adaptive augmentation, filtering, and refinement of training data, leading to more effective and efficient agentic post-training.

    \section{Task Formulation}\label{sec:task_formulation}
        This section presents the formulation of agentic post-training data curation.
Formally, let $\mathcal{D}_{\mathrm{raw}}$ denote a raw corpus of agentic post-training data. 
The task of data curation is to find a strategy $\rho$ that transforms the raw corpus into resources useful for training and inference:
\[
    \rho:
    \mathcal{D}_{\mathrm{raw}}
    \rightarrow
    \left(
    \mathcal{D}^{\mathrm{sft}}_{\rho},
    \mathcal{D}^{\mathrm{rl}}_{\rho},
    \mathcal{M}^{\mathrm{mem}}_{\rho}
    \right),
\]
where $\mathcal{D}^{\mathrm{sft}}_{\rho}$ is the curated SFT dataset, $\mathcal{D}^{\mathrm{rl}}_{\rho}$ is the curated RL dataset, and $\mathcal{M}^{\mathrm{mem}}_{\rho}$ is an inference-time memory bank.

Given the curated data produced by $\rho$, an agent policy is trained from a fixed base model $\pi_{\mathrm{base}}$ using a fixed training recipe:
\[
    \pi_{\rho}
    =
    \mathrm{Train}_{\mathrm{SFT+GRPO}}
    \left(
    \pi_{\mathrm{base}},
    \mathcal{D}^{\mathrm{sft}}_{\rho},
    \mathcal{D}^{\mathrm{rl}}_{\rho}
    \right).
\]
During inference, the policy can additionally condition on task-relevant memory retrieved from $\mathcal{M}^{\mathrm{mem}}_{\rho}$:
\[
    a_t \sim \pi_{\rho}
    \left(
    \cdot \mid q, o_{\leq t}, a_{<t},
    \mathrm{Retrieve}(q, \mathcal{M}^{\mathrm{mem}}_{\rho})
    \right),
\]
where $q$ is the user query, $o_{\leq t}$ are environment observations, and $a_{<t}$ are previous agent actions.

The objective is to identify a curation strategy whose processed data maximizes the downstream performance of the resulting agent. 
Using a held-out dev set $\mathcal{Q}_{\mathrm{dev}}$ as a proxy for generalization performance, the objective can be written as:
\[
    \rho^{\star}
    =
    \arg\max_{\rho}
    \;
    \mathrm{Perf}_{\mathrm{dev}}
    \left(
    \pi_{\rho},
    \mathcal{M}^{\mathrm{mem}}_{\rho}
    \right).
\]

    \section{Method}
        \begin{figure}[H]
    \centering
    \small
    \includegraphics[width=\linewidth]{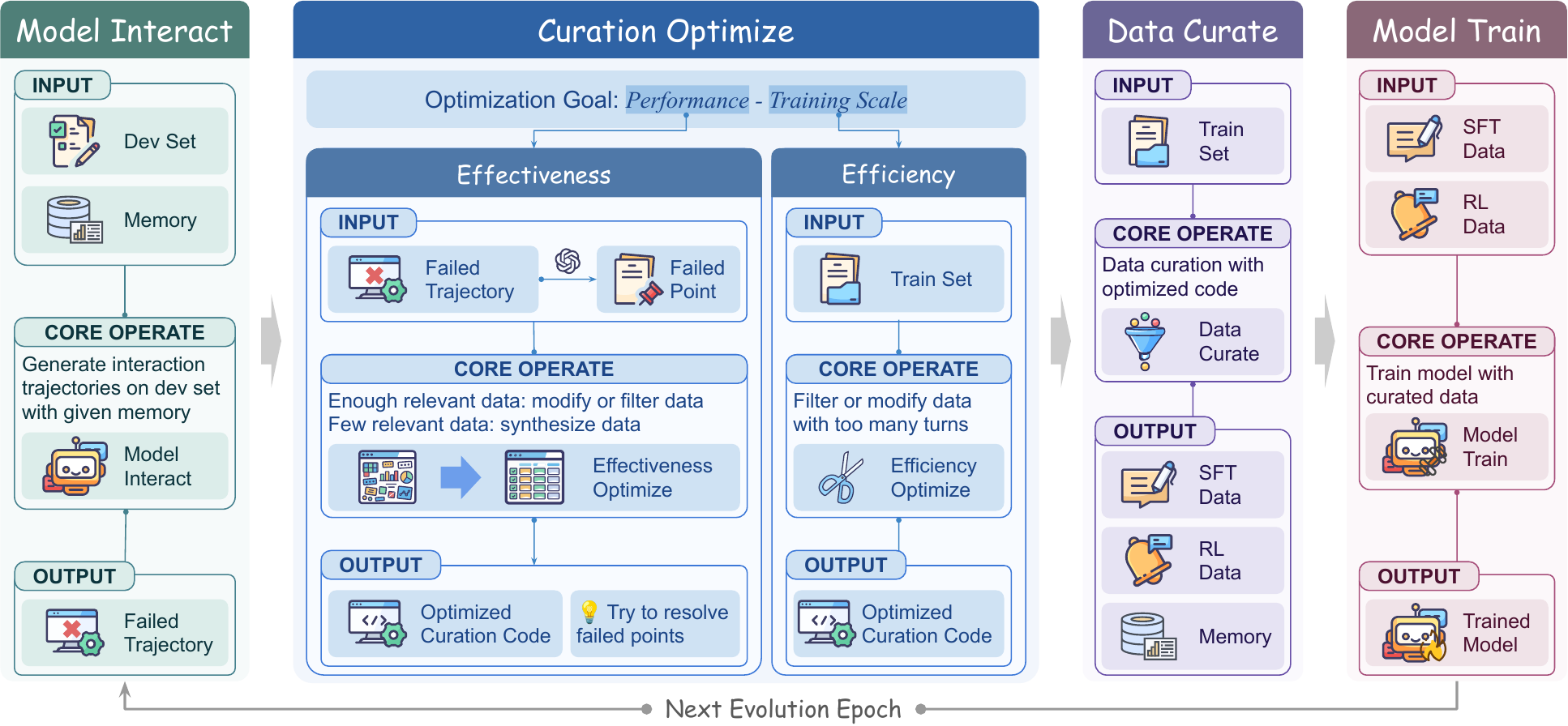}
    \caption{
        Overview of \ourmethod. 
        At evolution epoch $e$, the executable curation strategy $\rho_e$ transforms the fixed raw corpus into an SFT dataset, an RL dataset, and an inference-time memory bank. 
        Starting from the same base model, we train an agent using a fixed SFT+GRPO recipe and evaluate it on a held-out dev set. 
        An LLM-based code-evolution agent diagnoses recurring failure modes from failed dev trajectories, first revises $\rho_e$ through data augmentation, filtering, and refinement to improve effectiveness, and then removes and refines low-utility training turns to improve efficiency.
    }
    \label{fig:method}
\end{figure}

In this section, we propose \ourmethod, a failure-driven dynamic evolution framework for agentic post-training data curation. 
Following \S\ref{sec:task_formulation}, \ourmethod represents the curation strategy $\rho$ as executable code and directly evolves this code according to environment feedback.
Unlike methods that modify the agent policy, prompt, or interaction workflow, \ourmethod keeps the agentic post-training procedure fixed and optimizes how training and inference resources are constructed. 
The raw corpus $\mathcal{D}_{\mathrm{raw}}$ may contain human-annotated agent trajectories in the labeled setting or real user interaction trajectories in the wild setting.

Starting from an initial curation strategy $\rho_0$, \ourmethod performs multi-epoch evolution. At epoch $e$, the current strategy is executed on the fixed raw corpus:
\[
    \left(
        \mathcal{D}^{\mathrm{sft}}_e,
        \mathcal{D}^{\mathrm{rl}}_e,
        \mathcal{M}^{\mathrm{mem}}_e
    \right)
    \coloneqq
    \left(
        \mathcal{D}^{\mathrm{sft}}_{\rho_e},
        \mathcal{D}^{\mathrm{rl}}_{\rho_e},
        \mathcal{M}^{\mathrm{mem}}_{\rho_e}
    \right)
    =
    \rho_e\left(\mathcal{D}_{\mathrm{raw}}\right).
\]
We then restart agentic post-training from the same base model using the fixed SFT+GRPO recipe:
\[
    \pi_e
    \coloneqq
    \pi_{\rho_e}
    =
    \mathrm{Train}_{\mathrm{SFT+GRPO}}
    \left(
        \pi_{\mathrm{base}},
        \mathcal{D}^{\mathrm{sft}}_e,
        \mathcal{D}^{\mathrm{rl}}_e
    \right).
\]
During inference, $\pi_e$ additionally conditions on task-relevant memory:
\[
    a_t
    \sim
    \pi_e
    \left(
        \cdot
        \mid
        q,
        o_{\leq t},
        a_{<t},
        \mathrm{Retrieve}
        \left(q,\mathcal{M}^{\mathrm{mem}}_e\right)
    \right).
\]

The resulting agent is evaluated on a fixed held-out dev set. Failed interaction trajectories and statistics of the current curated data are then provided to an agent-based evolution, which revises $\rho_e$ to obtain $\rho_{e+1}$. The held-out test set is never used during this evolution process and is reserved for final evaluation.

\subsection{Feedback Collection}

    At each epoch, we evaluate $\left(\pi_e,\mathcal{M}^{\mathrm{mem}}_e\right)$ on the held-out dev set $\mathcal{Q}_{\mathrm{dev}}$. 
    For each task $q_i \in \mathcal{Q}_{\mathrm{dev}}$, the agent first retrieves task-relevant memory and then interacts with the environment until it terminates or reaches the interaction budget. We record the resulting trajectory as
    \[
        \tau_i^e
        =
        \left(
            q_i,
            m_i^e,
            \left\{
                \left(o_{i,t}^e,a_{i,t}^e\right)
            \right\}_{t=1}^{T_i^e},
            r_i^e,
            y_i^e,
            \phi_i^e
        \right),
    \]
    where
    \[
        m_i^e
        =
        \mathrm{Retrieve}
        \left(
            q_i,
            \mathcal{M}^{\mathrm{mem}}_e
        \right).
    \]
    Here, $o_{i,t}^e$ and $a_{i,t}^e$ denote the environment observation and agent action at turn $t$, respectively.
    $r_i^e$ is the benchmark reward or score, $y_i^e \in \{0,1\}$ is the final success indicator, and $\phi_i^e$ contains available auxiliary feedback, such as evaluator explanations, tool errors, or environment-side diagnostic messages.
    
    The failed trajectories used for curation evolution are
    \[
        \mathcal{T}^{-}_e
        =
        \left\{
            \tau_i^e
            \mid
            q_i \in \mathcal{Q}_{\mathrm{dev}},
            \ y_i^e = 0
        \right\}.
    \]
    In addition, we compute the current dev performance
    \[
        s_e
        =
        \mathrm{Perf}_{\mathrm{dev}}
        \left(
            \pi_e,
            \mathcal{M}^{\mathrm{mem}}_e
        \right)
    \]
    and collect a curation profile
    \[
        \Sigma_e
        =
        \mathrm{Stats}
        \left(
            \mathcal{D}^{\mathrm{sft}}_e,
            \mathcal{D}^{\mathrm{rl}}_e,
            \mathcal{M}^{\mathrm{mem}}_e
        \right),
    \]
    which summarizes properties such as data scale, interaction-turn distributions, trajectory lengths, and redundancy. Together, $\mathcal{T}^{-}_e$, $s_e$, and $\Sigma_e$ provide feedback about both agent effectiveness and data efficiency.
    
\subsection{Curation Evolution}
    The core of \ourmethod is an LLM-based code-evolution agent that directly rewrites the executable curation strategy. 
    Rather than treating effectiveness improvement and data reduction as two separate heuristics, we formulate curation evolution as a cost-aware optimization problem over curation strategies. 
    At epoch $e$, the evolution agent receives the current strategy $\rho_e$, the schema of $\mathcal{D}_{\mathrm{raw}}$, the failed dev trajectories $\mathcal{T}^{-}_e$, the dev performance $s_e$, and the curation profile $\Sigma_e$. It then searches for a revised strategy that improves downstream dev performance while controlling the amount of training supervision:
    \[
        \rho_{e+1}
        =
        \argmax_{\rho \in \mathcal{G}(\rho_e,\mathcal{T}^{-}_e,\Sigma_e)}
        \mathcal{J}(\rho),
    \]
    where $\mathcal{G}(\rho_e,\mathcal{T}^{-}_e,\Sigma_e)$ denotes the set of candidate code revisions proposed by the evolution agent. 
    The optimization objective is
    \[
        \mathcal{J}(\rho)
        =
        P(\rho)
        -
        \lambda C(\rho),
    \]
    where $P(\rho)$ measures the downstream effectiveness of the agent trained with the curated resources, $C(\rho)$ measures the normalized training-turn cost of the curated data, and $\lambda$ controls the effectiveness-efficiency trade-off. 
    In practice, this discrete code optimization is approximated by two ordered revisions, which first increase the effectiveness term $P(\rho)$ using failure-driven data curation, and then decrease the cost term $C(\rho)$ while preserving the newly covered supervision signals.
    Notably, if the evolved curation strategy in the current epoch does not improve the optimization objective, we roll it back to prevent error cascading.

    \subsubsection{Effectiveness Term.}
        The effectiveness term is computed by executing a candidate curation strategy, training the policy with the fixed recipe, and evaluating the resulting agent with its memory bank on the held-out dev set:
        \[
            P(\rho)
            =
            \mathrm{Perf}_{\mathrm{dev}}
            \left(
                \pi_{\rho},
                \mathcal{M}^{\mathrm{mem}}_{\rho}
            \right).
        \]
        Since fully training and evaluating every possible code revision is infeasible, \ourmethod uses failed dev trajectories from the current epoch as an actionable surrogate for improving this term. The evolution agent first diagnoses recurring failures:
        \[
            \mathcal{F}_e
            =
            g_{\mathrm{diag}}
            \left(
                \mathcal{T}^{-}_e
            \right),
        \]
        where each failure mode describes a missing or weak capability exposed by the current agent, such as incorrect tool selection, invalid argument grounding, weak multi-turn planning, poor recovery from tool errors, memory mismatch, or noisy supervision.
        We discussed the main failure modes discovered during evolving in Appendix~\ref{app:fail_mode_analysis}.
        Given these failure modes, the effectiveness-oriented revision updates the curation code as
        \[
            \widetilde{\rho}_{e+1}
            =
            g_{\mathrm{effect}}
            \left(
                \rho_e,
                \mathcal{F}_e,
                \Sigma_e
            \right).
        \]
        This revision optimizes $P(\rho)$ by changing how the fixed raw corpus is converted into SFT data, RL data, and memory. 
        Concretely, it may augment trajectories that cover uncovered failure modes with LLMs and simulated interactions, filter misleading or malformed examples, refine weak trajectories into clearer supervision, and extract memory entries that help with similar dev tasks at inference time. 
        Thus, the raw corpus remains fixed, but the curation code is rewritten to produce training and memory resources that better target the observed capability gaps.

    \subsubsection{Cost Term.}
        The cost term measures the scale of the training supervision produced by a curation strategy. After executing $\rho$ on the raw corpus, we compute the training-turn cost as
        \[
            \mathrm{Cost}_{\mathrm{train}}(\rho)
            =
            \sum_{z \in
            \mathcal{D}^{\mathrm{sft}}_{\rho}
            \cup
            \mathcal{D}^{\mathrm{rl}}_{\rho}}
            \ell(z),
        \]
        where $\ell(z)$ denotes the number of interaction turns retained in a curated training example. We then normalize the cost as
        \[
            C(\rho)
            =
            \tanh
            \left(
                \frac{\log(1+\mathrm{Cost}_{\mathrm{train}}(\rho))}{\log(1+N_{\mathrm{ref}})}
            \right),
        \]
        where $N_{\mathrm{ref}}$ is a reference constant that stabilizes the scale across datasets with long-tailed trajectory-length distributions.
        Based on this term, the efficiency-oriented revision further updates the intermediate strategy:
        \[
            \rho_{e+1}
            =
            g_{\mathrm{efficiency}}
            \left(
                \widetilde{\rho}_{e+1},
                \mathcal{F}_e,
                \Sigma_e,
                \lambda,
                N_{\mathrm{ref}}
            \right).
        \]
        This revision optimizes $C(\rho)$ by removing duplicate trajectories, pruning low-utility or noisy samples, compressing repeated interaction patterns, and truncating overly long trajectories. 
        To avoid harming $P(\rho)$, the pruning process is conditioned on the diagnosed failure modes and the curation profile, which samples that cover rare failures, preserve critical tool-use steps, or provide high-quality recovery behavior are retained, while redundant turns and weak supervision signals are preferentially removed. 
        As a result, the two revisions jointly approximate the objective $\mathcal{J}(\rho)$, where the first revision raises the expected dev performance, and the second revision reduces training scale under a performance-preserving constraint.

\subsection{Re-Curation and Fixed-Recipe Post-Training.}
    After obtaining $\rho_{e+1}$, we re-execute it on the fixed raw corpus:
    \[
    \left(
    \mathcal{D}^{\mathrm{sft}}_{e+1},
    \mathcal{D}^{\mathrm{rl}}_{e+1},
    \mathcal{M}^{\mathrm{mem}}_{e+1}
    \right)
    =
    \rho_{e+1}(\mathcal{D}_{\mathrm{raw}}).
    \]
    The next policy is trained from the same base model:
    \[
    \pi_{e+1}
    =
    \mathrm{Train}_{\mathrm{SFT+GRPO}}
    \left(
    \pi_{\mathrm{base}},
    \mathcal{D}^{\mathrm{sft}}_{e+1},
    \mathcal{D}^{\mathrm{rl}}_{e+1}
    \right).
    \]
    
    All other settings, including the base model, training recipe, environment, and inference budget, remain fixed across epochs. 
    The final strategy is selected based on the performance of the dev set, with the efficiency term used as a tie-breaker.

    \section{Experiment}
        \subsection{Experimental Setup}
    \paragraph{Raw Corpus and Benchmarks}
        We evaluate \ourmethod under the two raw-corpus settings: labeled data and wild data.
        \textit{Labeled data} consists of human-annotated agent trajectories from SWE-chat~\cite{baumann2026swechat}, AgentRewardBench~\cite{lu2025agentrewardbench}, and OpenHands-Feedback~\cite{openhands-feedback}.
        \textit{Wild data} consists of real user interaction trajectories from open sources, including ASSERT-KTH/reproducible-trajectories\footnote{\url{github.com/ASSERT-KTH/reproducible-trajectories}}, lelouch0110/claudeset-community\footnote{\url{huggingface.co/datasets/lelouch0110/claudeset-community}}, and nlile/misc-merged-claude-code-traces-v1\footnote{\url{huggingface.co/datasets/nlile/misc-merged-claude-code-traces-v1}}.
        For each setting, we build a fixed raw corpus $\mathcal{D}_{\mathrm{raw}}$ and split it into training and held-out development portions with a 9:1 ratio.
        The development portion serves as $\mathcal{Q}_{\mathrm{dev}}$ for evolving the executable curation strategy $\rho_e$.
        For final evaluation, we report held-out test performance on ACEBench-Agent~\cite{chen-etal-2025-acebench}, BFCL-V4~\cite{patil2025bfcl}, and $\tau^2$-Bench~\cite{barres2025tau2}, which are never used during evolution.
        More details of the benchmarks are provided in Appendix~\ref{app:experiment_benchmark}.
    \paragraph{Model and Baselines}
        Our experiments use \textsc{Qwen3-4B} as the base policy $\pi_{\mathrm{base}}$ and follow the fixed SFT+GRPO post-training recipe.
        We compare \ourmethod with GRPO without data curation and with representative agentic post-training data curation baselines, including MUA-RL~\cite{zhao2025mua}, EnvScaler~\cite{song2026envscaler}, AWM~\cite{wang2026agentworldmodelinfinity}, RODS~\cite{fang2026rodsrewarddrivenonlinedata}, and FunReason-MT~\cite{xu2025funreasonmttechnicalreportadvanced}.
        More details about these baselines are provided in Appendix~\ref{app:experiment_baseline}.
    \paragraph{Implementation Details}
        For each raw-corpus setting, \ourmethod evolves an independent curation strategy so that $\rho_e$ is adapted to the corresponding data distribution.
        The code-evolution agent uses \textsc{GPT-5.4} with mini-SWE-agent~\cite{yang2024sweagent} and runs for $3$ epochs.
        We use the cost-aware objective $\mathcal{J}(\rho)=P(\rho)-\lambda C(\rho)$ with $\lambda=0.3$ and $N_{\mathrm{ref}}=10^5$.
        During training, we use LoRA~\cite{hu2022lora} with rank $r=16$, \texttt{lora\_alpha}=32, and \texttt{lora\_dropout}=0.05.
        For SFT, we set the learning rate to $2.5\times10^{-5}$, the warmup ratio to $0.03$, use a cosine scheduler, and train for at most $140$ steps.
        For GRPO, we set the learning rate to $5\times10^{-6}$, the warmup ratio to $0.03$, $\beta_{\mathrm{KL}}=0.14$, the ratio clip to $0.05$, the advantage clip to $1.4$, and the maximum gradient norm to $1.0$.
        During inference, we set the temperature to $0.0$.
        All prompts used in our experiments are provided in Appendix~\ref{app:prompt}.
        The final evolved curation strategy is discussed in Appendix~\ref{app:evolved_curate_code}.

\subsection{Main Experiment}
    \begin{table}[H]
        \caption{
            Main experimental results.
            \ourmethod reports the mean and standard deviation over $3$ runs.
            The best result under each setting is marked in \textbf{bold}.
        }
        \label{tab:main_experiment}
        \centering
        \small
        \resizebox{\textwidth}{!}{\begin{tabular}{llcccccc}
    \toprule
    \multirow{2}{*}{\textbf{Method}} & \multirow{2}{*}{\textbf{Model}}
    & \multicolumn{3}{c}{\textbf{Labeled Data}}
    & \multicolumn{3}{c}{\textbf{Wild Data}} \\
    \cmidrule(lr){3-5} \cmidrule(l){6-8}
    &
    & \textbf{ACEBench}
    & \textbf{BFCL-V4}
    & \boldmath$\tau^2$
    & \textbf{ACEBench}
    & \textbf{BFCL-V4}
    & \boldmath$\tau^2$ \\
    
    \midrule
    \multicolumn{8}{@{}l}{\textit{GRPO w/o. Data Curation}} \\
    \addlinespace[0.2em]
    - & \textsc{Qwen3-4B}
        & $32.1$ & $13.8$ & $26.8$
        & $28.2$ & $12.5$ & $24.6$ \\
    - & \textsc{Qwen3-8B}
        & $36.8$ & $18.7$ & $31.8$
        & $30.9$ & $17.4$ & $29.6$ \\

    \midrule
    \multicolumn{8}{@{}l}{\textit{Prior Agent RL Training-Data Methods}} \\
    \addlinespace[0.2em]
    MUA-RL & \textsc{Qwen3-8B}
        & $53.3$ & $42.6$ & $32.8$
        & $50.7$ & $40.9$ & $30.6$ \\
    EnvScaler & \textsc{Qwen3-8B}
        & $50.4$ & $47.6$ & $37.9$
        & $48.8$ & $46.2$ & $35.5$ \\
    AWM & \textsc{Qwen3-8B}
        & $47.8$ & $39.7$ & $33.5$
        & $45.9$ & $38.4$ & $31.8$ \\
    RODS & \textsc{Qwen3-4B}
        & $48.3$ & $47.2$ & $31.6$
        & $46.2$ & $45.9$ & $29.6$ \\
    FunReason-MT & \textsc{Qwen3-4B}
        & $45.7$ & $50.3$ & $30.2$
        & $43.4$ & $48.7$ & $27.8$ \\

    \midrule
    \multicolumn{8}{@{}l}{\textit{Our Method}} \\
    \addlinespace[0.2em]
    \ourmethod & \textsc{Qwen3-4B}
        & $\bm{56.7 \pm 0.6}$ & $\bm{52.4 \pm 1.0}$ & $\bm{41.9 \pm 0.6}$
        & $\bm{55.8 \pm 1.4}$ & $\bm{50.0 \pm 0.6}$ & $\bm{37.2 \pm 1.2}$ \\
    \bottomrule
\end{tabular}}
    \end{table}

    Table~\ref{tab:main_experiment} shows that \ourmethod achieves the best performance across all benchmarks under both labeled and wild raw-corpus settings.
    Compared with the strongest prior result in each setting, \ourmethod improves the average score by $3.2$ on labeled data and $2.7$ on wild data.
    These gains support the central claim of this work, where evolving the executable curation strategy $\rho$ can yield better agentic post-training resources.
    Besides, from Table~\ref{tab:main_experiment}, we can also see that:
    \paragraph{Benchmark.}
        The improvements are consistent on ACEBench-Agent, BFCL-V4, and $\tau^2$-Bench, which stress different agent abilities.
        On ACEBench-Agent and BFCL-V4, \ourmethod improves tool selection, argument grounding, and format-following behavior.
        On $\tau^2$-Bench, where tasks require longer interaction and state tracking, \ourmethod also achieves the best result, suggesting that failure-driven curation can improve multi-turn decision making rather than only tool-call accuracy.
    \paragraph{Baseline.}
        GRPO without data curation performs substantially worse than curated-data methods, showing that raw trajectories provide weak training signals for agentic post-training.
        Prior baselines improve over raw GRPO but show different strengths across benchmarks and data settings.
        In contrast, \ourmethod consistently improves performance by revising $\rho_e$ according to held-out failures and by jointly curating $\mathcal{D}^{\mathrm{sft}}_{\rho}$, $\mathcal{D}^{\mathrm{rl}}_{\rho}$, and $\mathcal{M}^{\mathrm{mem}}_{\rho}$.
        Notably, \ourmethod with \textsc{Qwen3-4B} outperforms several prior methods using \textsc{Qwen3-8B}, indicating that adaptive curation can be as important as increasing model size.
    \paragraph{Raw Corpus.}
        Labeled data generally yields higher absolute scores than wild data because human-annotated trajectories are cleaner and more reliable.
        Nevertheless, \ourmethod remains the best method on wild data, demonstrating that the evolution process can extract useful supervision from noisy real interaction logs.
        This result matches the motivation of \ourmethod, where the raw corpus may be imperfect, but an adaptive curation strategy can select, refine, and organize it into more useful post-training and memory resources.

\subsection{Ablation Experiment}
    \begin{table}[H]
        \caption{
            Ablation study of \ourmethod.
            Operation ablations remove the effectiveness- or efficiency-oriented revision in curation evolution.
            Data ablations remove one curated resource.
        }
        \label{tab:ablation_study}
        \centering
        \small
        \setlength{\tabcolsep}{4.5pt}
\begin{tabular}{@{}lcccccc@{}}
    \toprule
    \multirow{2}{*}{\textbf{Method}} 
    & \multicolumn{3}{c}{\textbf{Labeled Data}}
    & \multicolumn{3}{c}{\textbf{Wild Data}} \\
    \cmidrule(lr){2-4}\cmidrule(l){5-7}
    & \textbf{ACE}
    & \textbf{BFCL}
    & \boldmath$\tau^2$
    & \textbf{ACE}
    & \textbf{BFCL}
    & \boldmath$\tau^2$ \\
    \midrule
    \ourmethod 
    & $56.7$ & $52.4$ & $41.9$ 
    & $55.8$ & $50.0$ & $37.2$ \\

    \midrule
    \multicolumn{7}{@{}l}{\textit{Operation Ablation}} \\
    \quad - Effectiveness 
    & $49.3$ & $44.4$ & $35.1$ 
    & $47.9$ & $42.4$ & $31.0$ \\
    \quad - Efficiency 
    & $55.5$ & $51.1$ & $40.8$ 
    & $54.3$ & $48.5$ & $35.9$ \\

    \midrule
    \multicolumn{7}{@{}l}{\textit{Data Ablation}} \\
    \quad - SFT Data 
    & $51.9$ & $45.2$ & $37.8$ 
    & $50.2$ & $43.5$ & $32.9$ \\
    \quad - RL Data 
    & $50.2$ & $47.6$ & $34.8$ 
    & $47.8$ & $44.6$ & $30.5$ \\
    \quad - Memory 
    & $53.9$ & $49.3$ & $39.6$ 
    & $52.2$ & $46.6$ & $34.7$ \\

    \bottomrule
\end{tabular}

    \end{table}

    Table~\ref{tab:ablation_study} validates the two parts of the objective $\mathcal{J}(\rho)=P(\rho)-\lambda C(\rho)$ and the three resources produced by $\rho$.
    Removing the effectiveness-oriented revision causes the largest degradation, with an average drop of about $7$ points across labeled and wild settings.
    This confirms that diagnosing failed trajectories and revising $\rho_e$ to cover the corresponding failure modes is the main driver of performance improvement.
    Removing the efficiency-oriented revision leads to a smaller but consistent drop, showing that reducing redundant or low-utility supervision can also improve data quality rather than only decrease the training scale.
    Data-level ablations further show that $\mathcal{D}^{\mathrm{sft}}_{\rho}$, $\mathcal{D}^{\mathrm{rl}}_{\rho}$, and $\mathcal{M}^{\mathrm{mem}}_{\rho}$ are complementary, where SFT and RL data mainly improve the policy, while memory provides reusable task-level knowledge at inference time.
    The drops are often larger under wild data, which indicates that noisy real trajectories depend more heavily on the adaptive filter and refinement.

\subsection{Efficiency of \ourmethod}
    \begin{figure}[!tbp]
        \centering
        \small
        \begin{tikzpicture}

\pgfplotsset{
efficiencyaxis/.style={
width=0.9\linewidth,
height=0.32\linewidth,
ymin=0,
symbolic x coords={mua,env,awm,rods,fun,ours},
xtick=data,
xticklabels={MUA,EnvS,AWM,RODS,FunR-MT,\ourmethod},
xticklabel style={font=\tiny, rotate=0, anchor=north},
tick label style={font=\tiny},
label style={font=\scriptsize},
ylabel style={xshift=0.2em},
title style={font=\scriptsize\bfseries},
axis line style={gray!55},
tick style={gray!55},
grid=both,
major grid style={gray!18},
minor grid style={gray!8},
enlarge x limits=0.11,
clip=false,
bar width=7.0pt,
}
}

\begin{scope}[xshift=0.080\linewidth]

\begin{axis}[
name=tokencost,
efficiencyaxis,
at={(0,0.48\linewidth)},
anchor=south west,
title={Token overhead / K turns},
ylabel={K tokens},
ymax=1520,
ytick={0,500,1000,1500},
]
\addplot+[
ybar,
draw=softgray!55,
fill=softgray!30,
fill opacity=0.78,
mark=none,
] coordinates {
(mua,1180) (env,940) (awm,720) (rods,820) (fun,1210)
};

\addplot+[
ybar,
draw=softred!85,
fill=softred!65,
fill opacity=0.86,
mark=none,
] coordinates {
(ours,510)
};

\node[font=\tiny, anchor=south, text=gray!70!black]
at ([yshift=1.2pt]axis cs:mua,1180) {1.18M};

\node[font=\tiny, anchor=south, text=gray!70!black]
at ([yshift=1.2pt]axis cs:env,940) {0.94M};

\node[font=\tiny, anchor=south, text=gray!70!black]
at ([yshift=1.2pt]axis cs:awm,720) {0.72M};

\node[font=\tiny, anchor=south, text=gray!70!black]
at ([yshift=1.2pt]axis cs:rods,820) {0.82M};

\node[font=\tiny, anchor=south, text=gray!70!black]
at ([yshift=1.2pt]axis cs:fun,1210) {1.21M};

\node[font=\tiny, anchor=south, text=softred!90!black]
at ([yshift=1.2pt]axis cs:ours,510) {0.51M};
\end{axis}

\begin{axis}[
name=timecost,
efficiencyaxis,
at={(0,0.08\linewidth)},
anchor=south west,
title={Wall-clock overhead / K turns},
ylabel={Seconds},
ymax=1360,
ytick={0,400,800,1200},
]
\addplot+[
ybar,
draw=softgray!55,
fill=softgray!30,
fill opacity=0.78,
mark=none,
] coordinates {
(mua,980) (env,760) (awm,590) (rods,690) (fun,1040)
};

\addplot+[
ybar,
draw=softred!85,
fill=softred!65,
fill opacity=0.86,
mark=none,
] coordinates {
(ours,405)
};

\node[font=\tiny, anchor=south, text=gray!70!black]
at ([yshift=1.2pt]axis cs:mua,980) {980s};

\node[font=\tiny, anchor=south, text=gray!70!black]
at ([yshift=1.2pt]axis cs:env,760) {760s};

\node[font=\tiny, anchor=south, text=gray!70!black]
at ([yshift=1.2pt]axis cs:awm,590) {590s};

\node[font=\tiny, anchor=south, text=gray!70!black]
at ([yshift=1.2pt]axis cs:rods,690) {690s};

\node[font=\tiny, anchor=south, text=gray!70!black]
at ([yshift=1.2pt]axis cs:fun,1040) {1040s};

\node[font=\tiny, anchor=south, text=softred!90!black]
at ([yshift=1.2pt]axis cs:ours,405) {405s};
\end{axis}

\node[
anchor=north,
inner sep=0pt,
font=\tiny
] at ($(timecost.south west)!0.5!(timecost.south east)+(0,-0.065\linewidth)$)
{
\begin{tabular}{@{}c@{\hspace{0.35em}}l@{\hspace{1.4em}}c@{\hspace{0.35em}}l@{}}

\tikz[baseline=-0.5ex]{
\filldraw[
draw=softgray!55,
fill=softgray!30,
fill opacity=0.78
] (0,0) rectangle (0.18,0.09);
}
&
Prior agent-RL data curation
&

\tikz[baseline=-0.5ex]{
\filldraw[
draw=softred!85,
fill=softred!65,
fill opacity=0.86
] (0,0) rectangle (0.18,0.09);
}
&
\ourmethod

\end{tabular}
};

\end{scope}

\end{tikzpicture}
        \caption{
            Averaged token and wall-clock overhead of each curation method per $1K$ retained training turns on three benchmarks.
        }
        \label{fig:efficiency}
    \end{figure}
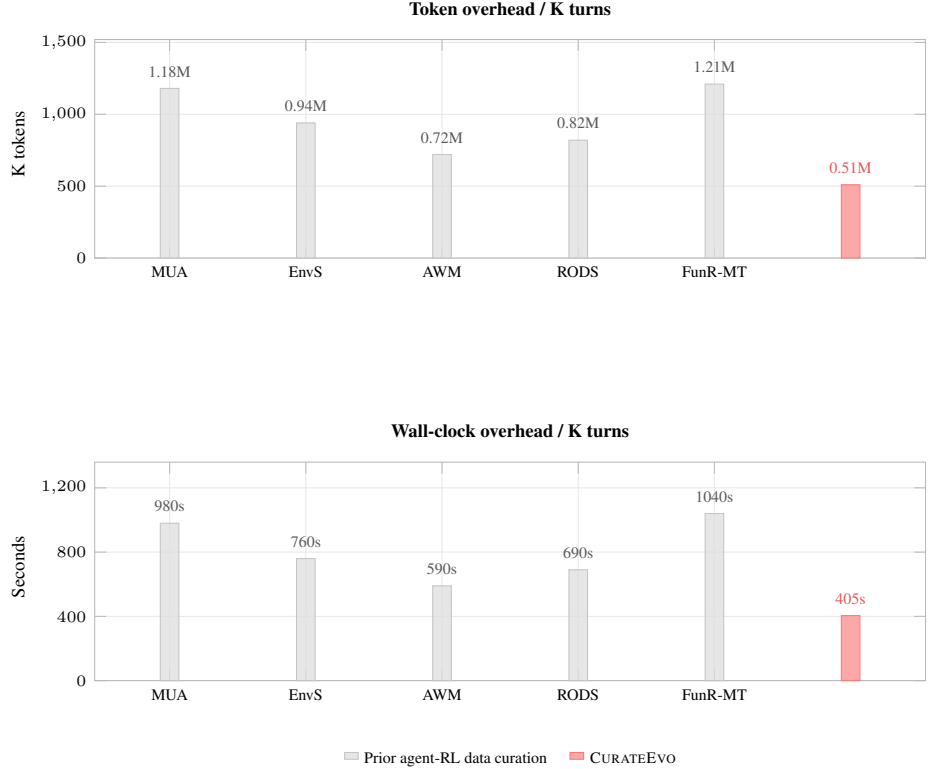

    Figure~\ref{fig:efficiency} compares the overhead of executing different data-curation methods.
    \ourmethod requires $0.51$M tokens and $405$ seconds per $1K$ retained training turns, reducing token overhead by about $48\%$ and wall-clock overhead by about $50\%$ relative to the average prior baseline.
    This efficiency comes from evolving curation code over the fixed raw corpus instead of relying on extra environment interactions, simulated users, or heavy auxiliary data generation.
    Together with Table~\ref{tab:main_experiment}, the result shows that cost-aware curation can improve final performance while keeping the data-curation budget practical.
    It also complements the cost term $C(\rho)$, where \ourmethod reduces not only the retained training scale but also the overhead needed to construct the curated resources.

\subsection{Impact of Different Factors on \ourmethod}
    \subsubsection{Agentic Post-Training Recipe}
        \begin{table}[H]
            \caption{
                Compatibility with different agentic post-training recipes on \textsc{Qwen3-4B}.
                ACE denotes ACEBench and BFCL denotes BFCL-V4.
            }
            \label{tab:performace_agent_rl}
            \centering
            \small
            \setlength{\tabcolsep}{5pt}
\begin{tabular}{@{}lcccccc@{}}
    \toprule
    \multirow{2}{*}{\textbf{Method}} 
    & \multicolumn{3}{c}{\textbf{Labeled Data}}
    & \multicolumn{3}{c}{\textbf{Wild Data}} \\
    \cmidrule(lr){2-4}\cmidrule(l){5-7}
    & \textbf{ACE}
    & \textbf{BFCL}
    & \boldmath$\tau^2$
    & \textbf{ACE}
    & \textbf{BFCL}
    & \boldmath$\tau^2$ \\
    \midrule
    GRPO & $32.1$ & $13.8$ & $26.8$ & $28.2$ & $12.5$ & $24.6$ \\
    \quad w. \ourmethod & $56.7$ & $52.4$ & $41.9$ & $55.8$ & $50.0$ & $37.2$ \\
    \midrule
    AgentGym-RL & $42.6$ & $18.4$ & $34.2$ & $37.5$ & $16.9$ & $31.5$ \\
    \quad w. \ourmethod & $57.8$ & $53.5$ & $42.7$ & $56.9$ & $50.9$ & $38.1$ \\
    \midrule
    ProRL-Agent & $45.9$ & $19.8$ & $37.8$ & $40.2$ & $17.9$ & $34.8$ \\
    \quad w. \ourmethod & $58.7$ & $54.4$ & $43.4$ & $57.6$ & $51.7$ & $38.8$ \\
    \bottomrule
\end{tabular}

        \end{table}

        The main experiments keep the SFT+GRPO recipe fixed during evolution.
        Table~\ref{tab:performace_agent_rl} further tests whether the curated resources produced by the final strategy can benefit other post-training recipes.
        Across GRPO, AgentGym-RL, and ProRL-Agent, adding \ourmethod improves all benchmarks and both raw-corpus settings, with an average gain of $21.3$ points.
        The strongest results are obtained when \ourmethod is combined with ProRL-Agent, showing that stronger policy optimization and better data curation are complementary.
        Thus, \ourmethod should be viewed as a data-side module, which improves $\mathcal{D}^{\mathrm{sft}}_{\rho}$, $\mathcal{D}^{\mathrm{rl}}_{\rho}$, and $\mathcal{M}^{\mathrm{mem}}_{\rho}$, while remaining compatible with different agentic post-training recipes.

    \subsubsection{Evolution Epoch}
        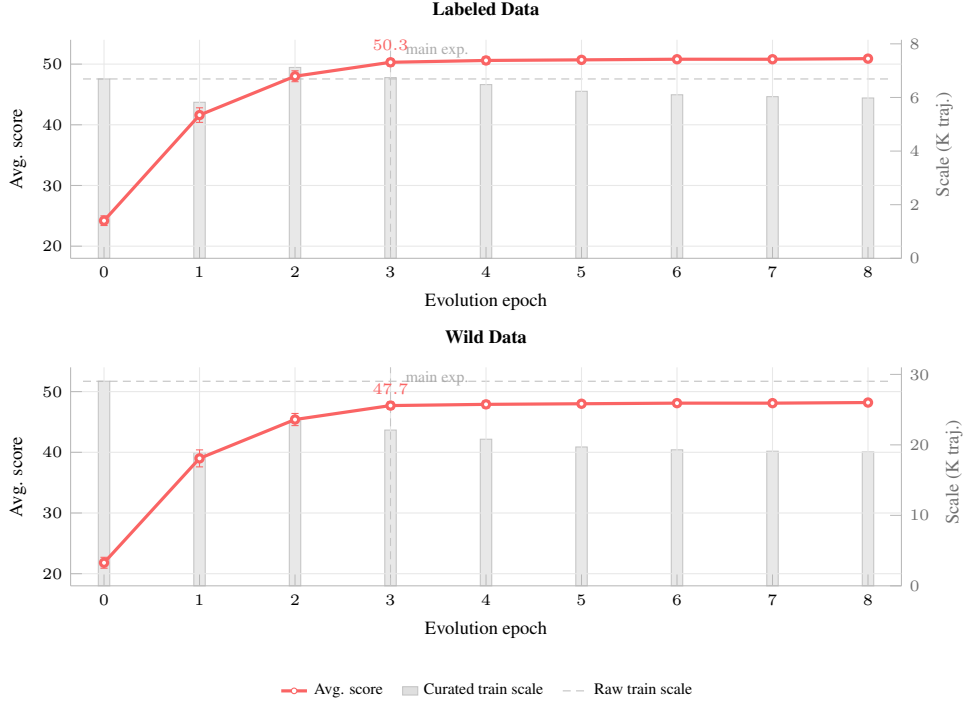
\begin{figure}[H]
            \centering
            \small
            \begin{tikzpicture}

\pgfplotsset{
evolveaxis/.style={
width=0.9\linewidth,
height=0.32\linewidth,
xmin=-0.35,
xmax=8.35,
xtick={0,1,2,3,4,5,6,7,8},
tick label style={font=\tiny},
label style={font=\scriptsize},
title style={font=\scriptsize\bfseries},
grid=both,
major grid style={gray!18},
minor grid style={gray!8},
axis line style={gray!55},
tick style={gray!55},
clip=false,
}
}

\begin{scope}[xshift=0.08\linewidth]

\begin{axis}[
name=labeledscale,
evolveaxis,
at={(0,0.45\linewidth)},
anchor=south west,
ymin=0,
ymax=8.15,
ytick={0,2,4,6,8},
ylabel={Scale (K traj.)},
ylabel style={softgray!75!black, xshift=0.2em},
yticklabel style={softgray!75!black},
axis y line*=right,
axis x line=none,
grid=none,
axis background/.style={fill=none},
]
\addplot+[
ybar,
bar width=4.3pt,
draw=softgray!55,
fill=softgray!30,
fill opacity=0.72,
mark=none,
forget plot,
] coordinates {
(0,6.69) (1,5.82) (2,7.12) (3,6.74) (4,6.48)
(5,6.23) (6,6.10) (7,6.03) (8,5.98)
};

\addplot+[
softgray!65,
densely dashed,
very thin,
no marks,
forget plot
] coordinates {
(-0.22,6.69) (8.22,6.69)
};
\end{axis}

\begin{axis}[
name=labeledperf,
evolveaxis,
at={(labeledscale.south west)},
anchor=south west,
title={Labeled Data},
ymin=18,
ymax=54.0,
ytick={20,30,40,50},
xlabel={Evolution epoch},
ylabel={Avg. score},
axis y line*=left,
axis x line*=bottom,
axis background/.style={fill=none},
]
\addplot+[
gray!45,
densely dashed,
no marks,
forget plot
] coordinates {
(3,18) (3,54.0)
};

\node[
anchor=west,
font=\tiny,
text=gray!65
] at (axis cs:3.07,52.25) {main exp.};

\addplot+[
softred,
very thick,
mark=*,
mark size=1.35pt,
mark options={fill=white, draw=softred},
error bars/.cd,
y dir=both,
y explicit,
] coordinates {
(0,24.2) +- (0,0.8)
(1,41.6) +- (0,1.2)
(2,48.0) +- (0,0.9)
(3,50.3) +- (0,0.6)
(4,50.6) +- (0,0.6)
(5,50.7) +- (0,0.5)
(6,50.8) +- (0,0.5)
(7,50.8) +- (0,0.5)
(8,50.9) +- (0,0.5)
};

\node[
anchor=south,
font=\tiny,
text=softred
] at (axis cs:3,50.85) {$50.3$};
\end{axis}

\begin{axis}[
name=wildscale,
evolveaxis,
at={(0,0.14\linewidth)},
anchor=south west,
ymin=0,
ymax=31.0,
ytick={0,10,20,30},
ylabel={Scale (K traj.)},
ylabel style={softgray!75!black, xshift=0.2em},
yticklabel style={softgray!75!black},
axis y line*=right,
axis x line=none,
grid=none,
axis background/.style={fill=none},
]
\addplot+[
ybar,
bar width=4.3pt,
draw=softgray!55,
fill=softgray!30,
fill opacity=0.72,
mark=none,
forget plot,
] coordinates {
(0,29.03) (1,18.9) (2,23.4) (3,22.1) (4,20.8)
(5,19.7) (6,19.3) (7,19.1) (8,19.0)
};

\addplot+[
softgray!65,
densely dashed,
very thin,
no marks,
forget plot
] coordinates {
(-0.22,29.03) (8.22,29.03)
};
\end{axis}

\begin{axis}[
name=wildperf,
evolveaxis,
at={(wildscale.south west)},
anchor=south west,
title={Wild Data},
ymin=18,
ymax=54.0,
ytick={20,30,40,50},
xlabel={Evolution epoch},
ylabel={Avg. score},
axis y line*=left,
axis x line*=bottom,
axis background/.style={fill=none},
]
\addplot+[
gray!45,
densely dashed,
no marks,
forget plot
] coordinates {
(3,18) (3,54.0)
};

\node[
anchor=west,
font=\tiny,
text=gray!65
] at (axis cs:3.07,52.25) {main exp.};

\addplot+[
softred,
very thick,
mark=*,
mark size=1.35pt,
mark options={fill=white, draw=softred},
error bars/.cd,
y dir=both,
y explicit,
] coordinates {
(0,21.8) +- (0,0.9)
(1,39.0) +- (0,1.4)
(2,45.4) +- (0,1.0)
(3,47.7) +- (0,0.6)
(4,47.9) +- (0,0.6)
(5,48.0) +- (0,0.5)
(6,48.1) +- (0,0.5)
(7,48.1) +- (0,0.5)
(8,48.2) +- (0,0.5)
};

\node[
anchor=south,
font=\tiny,
text=softred
] at (axis cs:3,48.25) {$47.7$};
\end{axis}

\node[
anchor=north,
inner sep=0pt,
font=\tiny
] at ($(wildperf.south west)!0.5!(wildperf.south east)+(0,-0.09\linewidth)$)
{
\begin{tabular}{@{}c@{\hspace{0.35em}}l@{\hspace{1.2em}}c@{\hspace{0.35em}}l@{\hspace{1.2em}}c@{\hspace{0.35em}}l@{}}

\tikz[baseline=-0.5ex]{
\draw[softred, very thick] (0,0) -- (0.34,0);
\filldraw[draw=softred, fill=white] (0.17,0) circle (1.05pt);
}
&
Avg. score
&

\tikz[baseline=-0.5ex]{
\filldraw[draw=softgray!55, fill=softgray!30] (0,-0.055) rectangle (0.20,0.055);
}
&
Curated train scale
&

\tikz[baseline=-0.5ex]{
\draw[softgray!65, densely dashed, very thin] (0,0) -- (0.34,0);
}
&
Raw train scale

\end{tabular}
};

\end{scope}

\end{tikzpicture}
            \caption{
                Averaged performance and curated training scale across evolution epochs on three benchmarks.
            }
            \label{fig:evolve_epoch}
        \end{figure}

        Figure~\ref{fig:evolve_epoch} reports dev performance and curated training scale as the curation strategy evolves from $\rho_0$. This scale is an observable proxy for the training-turn cost $C(\rho)$.
        Most gains appear in the first three epochs, where the average score increases from $24.2$ to $50.3$ on labeled data and from $21.8$ to $47.7$ on wild data.
        After the epoch used in the main experiments, performance becomes nearly saturated, while the curated scale continues to decrease slightly.
        This trend supports the design of failure-driven evolution, where early revisions mainly increase $P(\rho)$ by covering missing capabilities, and later revisions primarily reduce $C(\rho)$ by removing redundant supervision.
        The curated scale does not need to grow monotonically with performance, which indicates that the goal of curation is to construct a more useful training distribution rather than a larger one.
    
    \subsubsection{\texorpdfstring{$\lambda$}{lambda} in the Evolution Objective}
        \begin{figure}[!tbp]
            \centering
            \small
            \begin{tikzpicture}

\pgfplotsset{
lambdaaxis/.style={
width=0.9\linewidth,
height=0.32\linewidth,
xmin=-0.045,
xmax=1.045,
xtick={0,0.1,0.2,0.3,0.4,0.5,0.6,0.7,0.8,0.9,1.0},
xticklabels={0.0,0.1,0.2,0.3,0.4,0.5,0.6,0.7,0.8,0.9,1.0},
xticklabel style={font=\tiny, rotate=45, anchor=east},
tick label style={font=\tiny},
label style={font=\scriptsize},
title style={font=\scriptsize\bfseries},
grid=both,
major grid style={gray!18},
minor grid style={gray!8},
axis line style={gray!55},
tick style={gray!55},
clip=false,
}
}
\begin{scope}[xshift=0.08\linewidth]

\begin{axis}[
name=labeledlambdascale,
lambdaaxis,
at={(0,0.45\linewidth)},
anchor=south west,
ymin=0,
ymax=8.9,
ytick={0,2,4,6,8},
ylabel={Scale (K traj.)},
ylabel style={softgray!75!black, xshift=0.2em},
yticklabel style={softgray!75!black},
axis y line*=right,
axis x line=none,
xtick=\empty,
grid=none,
axis background/.style={fill=none},
]
\addplot+[
ybar,
bar width=3.2pt,
draw=softgray!55,
fill=softgray!30,
fill opacity=0.72,
mark=none,
forget plot,
] coordinates {
(0.0,8.21) (0.1,7.46) (0.2,6.81) (0.3,6.23)
(0.4,5.92) (0.5,5.48) (0.6,4.97) (0.7,4.42)
(0.8,3.98) (0.9,3.61) (1.0,3.27)
};

\addplot+[
softgray!65,
densely dashed,
very thin,
no marks,
forget plot
] coordinates {
(-0.03,6.69) (1.03,6.69)
};

\node[
anchor=north,
font=\tiny,
text=softgray!75!black
] at (axis cs:0.3,5.86) {6.23K};
\end{axis}

\begin{axis}[
name=labeledlambda,
lambdaaxis,
at={(labeledlambdascale.south west)},
anchor=south west,
title={Labeled Data},
ymin=40,
ymax=52.5,
ytick={40,44,48,52},
xlabel={$\lambda$},
ylabel={Avg. score},
axis y line*=left,
axis x line*=bottom,
axis background/.style={fill=none},
]
\addplot+[
gray!45,
densely dashed,
no marks,
forget plot
] coordinates {
(0.3,40) (0.3,52.5)
};

\node[
anchor=west,
font=\tiny,
text=gray!65
] at (axis cs:0.315,51.4) {main exp.};

\addplot+[
softred,
very thick,
mark=*,
mark size=1.25pt,
mark options={fill=white, draw=softred},
error bars/.cd,
y dir=both,
y explicit,
] coordinates {
(0.0,49.0) +- (0,1.0)
(0.1,49.7) +- (0,0.9)
(0.2,50.1) +- (0,0.7)
(0.3,50.3) +- (0,0.6)
(0.4,50.0) +- (0,0.7)
(0.5,49.4) +- (0,0.8)
(0.6,48.5) +- (0,0.9)
(0.7,47.6) +- (0,1.0)
(0.8,46.3) +- (0,1.1)
(0.9,45.3) +- (0,1.2)
(1.0,44.1) +- (0,1.3)
};

\node[
anchor=south,
font=\tiny,
text=softred
] at (axis cs:0.3,50.72) {$50.3$};
\end{axis}

\begin{axis}[
name=wildlambdascale,
lambdaaxis,
at={(0,0.14\linewidth)},
anchor=south west,
ymin=0,
ymax=36.5,
ytick={0,10,20,30},
ylabel={Scale (K traj.)},
ylabel style={softgray!75!black, xshift=0.2em},
yticklabel style={softgray!75!black},
axis y line*=right,
axis x line=none,
xtick=\empty,
grid=none,
axis background/.style={fill=none},
]
\addplot+[
ybar,
bar width=3.2pt,
draw=softgray!55,
fill=softgray!30,
fill opacity=0.72,
mark=none,
forget plot,
] coordinates {
(0.0,33.8) (0.1,28.6) (0.2,23.5) (0.3,19.7)
(0.4,17.9) (0.5,15.6) (0.6,13.4) (0.7,12.0)
(0.8,10.4) (0.9,9.5) (1.0,8.6)
};

\addplot+[
softgray!65,
densely dashed,
very thin,
no marks,
forget plot
] coordinates {
(-0.03,29.03) (1.03,29.03)
};

\node[
anchor=north,
font=\tiny,
text=softgray!75!black
] at (axis cs:0.3,18.95) {19.7K};
\end{axis}

\begin{axis}[
name=wildlambda,
lambdaaxis,
at={(wildlambdascale.south west)},
anchor=south west,
title={Wild Data},
ymin=40,
ymax=52.5,
ytick={40,44,48,52},
xlabel={$\lambda$},
ylabel={Avg. score},
axis y line*=left,
axis x line*=bottom,
axis background/.style={fill=none},
]
\addplot+[
gray!45,
densely dashed,
no marks,
forget plot
] coordinates {
(0.3,40) (0.3,52.5)
};

\node[
anchor=west,
font=\tiny,
text=gray!65
] at (axis cs:0.315,51.4) {main exp.};

\addplot+[
softred,
very thick,
mark=*,
mark size=1.25pt,
mark options={fill=white, draw=softred},
error bars/.cd,
y dir=both,
y explicit,
] coordinates {
(0.0,46.0) +- (0,1.2)
(0.1,46.9) +- (0,1.0)
(0.2,47.4) +- (0,0.8)
(0.3,47.7) +- (0,0.6)
(0.4,47.3) +- (0,0.7)
(0.5,46.6) +- (0,0.8)
(0.6,45.8) +- (0,0.9)
(0.7,44.8) +- (0,1.0)
(0.8,43.5) +- (0,1.1)
(0.9,42.3) +- (0,1.2)
(1.0,41.0) +- (0,1.3)
};

\node[
anchor=south,
font=\tiny,
text=softred
] at (axis cs:0.3,48.12) {$47.7$};
\end{axis}

\node[
anchor=north,
inner sep=0pt,
font=\tiny
] at ($(wildlambda.south west)!0.5!(wildlambda.south east)+(0,-0.1\linewidth)$)
{
\begin{tabular}{@{}c@{\hspace{0.35em}}l@{\hspace{1.2em}}c@{\hspace{0.35em}}l@{\hspace{1.2em}}c@{\hspace{0.35em}}l@{}}

\tikz[baseline=-0.5ex]{
\draw[softred, very thick] (0,0) -- (0.34,0);
\filldraw[draw=softred, fill=white] (0.17,0) circle (1.0pt);
}
&
Avg. score
&

\tikz[baseline=-0.5ex]{
\filldraw[draw=softgray!55, fill=softgray!30] (0,-0.055) rectangle (0.20,0.055);
}
&
Curated train scale
&

\tikz[baseline=-0.5ex]{
\draw[softgray!65, densely dashed, very thin] (0,0) -- (0.34,0);
}
&
Raw train scale

\end{tabular}
};

\end{scope}

\end{tikzpicture}
            \caption{
                Effect of the cost weight $\lambda$ on averaged performance and curated training scale on three benchmarks.
            }
            \label{fig:lambda_effect}
        \end{figure}
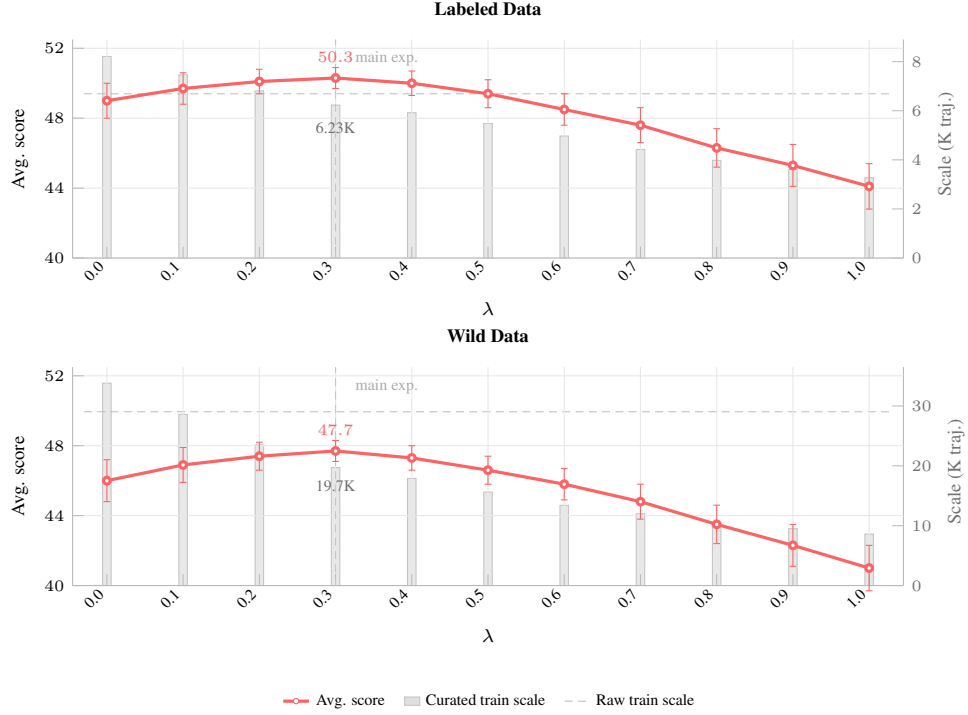

        Figure~\ref{fig:lambda_effect} studies the cost weight $\lambda$ in our optimized objective $\mathcal{J}(\rho)$.
        When $\lambda=0$, the evolution process focuses on dev performance and keeps a larger curated set, but this does not give the best result.
        As $\lambda$ increases, the curated training scale steadily decreases.
        A moderate value, $\lambda=0.3$, achieves the best average performance for both labeled and wild data and is therefore used in the main experiments.
        When $\lambda$ is too large, the score drops because over-pruning can remove useful long-horizon trajectories or rare failure cases.
        These results show that the effectiveness term $P(\rho)$ and the cost term $C(\rho)$ should be optimized jointly, rather than treating efficiency as a separate post-processing step.

    \section{Conclusion}
        In this work, we introduced \ourmethod, a failure-driven dynamic evolution framework for agentic post-training data curation. Instead of relying on a fixed curation pipeline, \ourmethod represents the curation strategy as executable code and iteratively improves it using failed trajectories from a held-out development set. By jointly constructing SFT data, RL data, and an inference-time memory bank, the framework adapts the raw corpus to the weaknesses exposed by the current agent. Its cost-aware evolution objective further balances effectiveness and efficiency by first targeting recurring failure modes and then reducing redundant or low-utility training turns.
        Experiments across ACEBench-Agent, BFCL-V4, and $\tau^2$-Bench show that \ourmethod achieves consistent improvements under both labeled and wild-data settings, outperforming prior data-curation baselines by $3.2$ and $2.7$ average points, respectively. Ablation studies confirm that both effectiveness-oriented and efficiency-oriented revisions contribute to the final performance, and that SFT data, RL data, and memory provide complementary benefits. Additional analyses show that the curated resources produced by \ourmethod can be combined with different agentic post-training recipes and can reduce curation overhead while maintaining strong downstream performance. Overall, these results suggest that adaptive, feedback-driven data curation is a practical and effective direction for improving LLM agents.

    \clearpage
    \bibliographystyle{plainnat}
    \bibliography{aaai2027}

    \clearpage
    \appendix
    \section{Prompt}\label{app:prompt}
    \begin{center}
        \small
        \begin{prompt}{Failed Point Generation from Failed Trajectories}
    \raggedright
    \setlength{\parindent}{0pt}
    \setlength{\parskip}{0.45em}
    \linespread{1.08}\selectfont
    
    You are a failure analysis agent for improving data curation in agent RL. Given a failed trajectory, identify the key failed points that reveal why the current training data is insufficient.
    
    A \textbf{failed point} is a concrete and actionable weakness exposed by the failed trajectory. It should describe what capability or behavior failed, where the failure is reflected in the trajectory, and how the data curation code can be improved to address it. A failed point should not simply restate that the agent failed the task.
    
    \textbf{Input.}
    
    \textbf{Task:}
    
    \texttt{\{TASK\}}
    
    \textbf{Failed trajectory:}
    
    \texttt{\{FAILED\_TRAJECTORY\}}
    
    \textbf{Feedback:}
    
    \texttt{\{FEEDBACK\}}
    
    \textbf{Generation rules.}
    Focus only on failures that can guide data selection, filtering, rewriting, or memory construction. Do not summarize the whole trajectory. Do not include irrelevant details. Do not output generic issues such as ``the agent failed to solve the task.'' Merge duplicate or highly similar issues. Prefer concise, specific, and actionable failed points.
    
    \textbf{Procedure.}
    Read the task, failed trajectory, and feedback carefully. Identify the decisive failure step or behavior. Infer the underlying capability weakness exposed by the failure. Then describe how the data curation code should be improved to address this weakness.
    
    Think step by step internally, but do not show your reasoning. Output 1 to 3 failed points in a JSON array. Each failed point must contain exactly the following fields:
    
    \texttt{failed\_point}: the capability or behavior that failed.
    
    \texttt{evidence}: the specific part of the trajectory or feedback that shows this failure.
    
    \texttt{curation\_hint}: how the data curation code should be improved to address it.
    
    \textbf{Output format:}
    
    \texttt{[}
    
    \texttt{\ \ \{}
    
    \texttt{\ \ \ \ \ "failed\_point": "...",}
    
    \texttt{\ \ \ \ \ "evidence": "...",}
    
    \texttt{\ \ \ \ \ "curation\_hint": "..."}
    
    \texttt{\ \ \}}
    
    \texttt{]}
\end{prompt}

        \captionof{figure}{
            The prompt to transfer failed trajectories into failed points.
        }
        \label{fig:failed_point_prompt}
    \end{center}
    \begin{center}
        \small
        \begin{prompt}{Failure-Guided Data-Curation Evolution}
    \raggedright
    \setlength{\parindent}{0pt}
    \setlength{\parskip}{0.45em}
    \linespread{1.08}\selectfont
    
    You are an autonomous software engineering agent working inside the SWE-agent harness. Your task is to modify the repository's data curation code so that it can automatically read failed points, inspect the current training data, diagnose failure modes, and improve the curation logic accordingly.
    
    \textbf{Input.}
    
    \textbf{Failed points path:}
    
    \texttt{\{FAILED\_POINTS\_PATH\}}
    
    \textbf{Training data path:}
    
    \texttt{\{TRAIN\_DATA\_PATH\}}
    
    \textbf{Curation code or search root:}
    
    \texttt{\{CURATION\_CODE\_PATH\}}
    
    \textbf{Output path:}
    
    \texttt{\{OUTPUT\_PATH\}}
    
    \textbf{Optional validation command:}
    
    \texttt{\{VALIDATION\_COMMAND\}}
    
    \textbf{Cost weight:}
    
    \texttt{\{LAMBDA\}}, default to \texttt{0.3} if not specified.
    
    \textbf{Task requirements.}
    First locate the existing data curation pipeline and preserve its original interfaces whenever possible. Then add or improve support for loading failed points from \texttt{\{FAILED\_POINTS\_PATH\}} and reading the current training data from \texttt{\{TRAIN\_DATA\_PATH\}}. Infer schemas from real samples rather than assuming fixed fields.
    
    Use the failed points to identify recurring failure modes, such as missing supervision, incorrect tool-use format, wrong API arguments, poor long-horizon planning, weak error recovery, memory mismatch, noisy trajectories, redundant trajectories, or inconsistent labels.
    
    Modify the curation code so these failure modes guide data selection, filtering, rewriting, deduplication, trajectory truncation, and memory construction if the repository supports memory. Do not simply increase the dataset size. The updated pipeline should remain deterministic and re-runnable from raw training data.
    
    Use cost-aware selection. Prefer examples that cover uncovered failures and penalize long, redundant, noisy, or malformed trajectories. Use \texttt{\{LAMBDA\}} as the cost weight when available.
    
    \textbf{Constraints.}
    Do not manually edit raw training data, failed-point files, dev/test sets, evaluator logic, or tests. Do not add heavy dependencies unless already used by the project. Keep the patch minimal, auditable, and backward compatible.
    
    \textbf{Validation.}
    Run syntax checks on modified Python files. Run relevant tests if available. Run a lightweight dry run of the curation pipeline with the given failed-points and training-data paths. If \texttt{\{VALIDATION\_COMMAND\}} is provided and feasible, run it.
    
    \textbf{Final response.}
    Summarize the files changed, how failed points are loaded, how training data is inspected, how failure modes affect curation, how cost and scale are controlled, which validation commands were run, and any remaining assumptions or limitations.
\end{prompt}

        \captionof{figure}{
            The prompt to optimize curation code based on failed points and existing training data.
        }
        \label{fig:curation_evolve_prompt}
    \end{center}

    The prompts used by \ourmethod are shown in Figure~\ref{fig:failed_point_prompt} and Figure~\ref{fig:curation_evolve_prompt}.

\section{Experimental Resource}\label{app:experiment_resource}
    \subsection{Benchmark}\label{app:experiment_benchmark}
        \paragraph{ACEBench.}
            ACEBench is a tool-use benchmark for evaluating whether LLM agents can select appropriate tools, fill arguments, and complete tasks under realistic instructions~\cite{chen-etal-2025-acebench}. It contains three evaluation categories: Normal, Special, and Agent. Normal covers basic function-calling cases; Special stresses imperfect instructions with ambiguity or incompleteness; Agent evaluates multi-turn tool use through multi-agent interactions. In our experiments, ACEBench is used as a held-out test of fine-grained tool invocation and robustness to instruction uncertainty. It is particularly relevant to data curation because errors can expose whether the curated trajectories teach reliable tool schemas, argument grounding, and recovery from under-specified user requests.
        \paragraph{BFCL-V4.}
            The Berkeley Function Calling Leaderboard V4 evaluates tool/function calling accuracy for LLMs and expands earlier static function calling into more agentic settings~\cite{patil2025bfcl}. It includes categories such as non-live and live calls, multi-turn interaction, hallucination measurement, and agentic tasks. V4 further introduces web search, memory, and format-sensitivity tests. We use BFCL-V4 as a held-out benchmark for standardized real-world tool invocation: the agent must map natural-language goals to valid calls, maintain state across turns, avoid unsupported calls, and produce outputs following required schemas. This complements ACEBench by emphasizing broad function-calling coverage and executable correctness, making it useful for testing whether the evolved curation code improves transferable tool-use behavior rather than overfitting to a single environment.
        \paragraph{$\tau^2$-Bench.}
            $\tau^2$-Bench evaluates conversational agents in a dual-control environment~\cite{barres2025tau2}. Unlike single-control benchmarks, where only the agent changes the world state, $\tau^2$-Bench places the agent and a simulated user in a shared Telecom environment; both sides have distinct tools and partial observations. Tasks are generated compositionally and are verifiable, requiring the agent not only to reason and call tools but also to coordinate with and guide the user. We use it as the long-horizon interaction benchmark because success depends on communication, state tracking, tool use, and policy following across multiple turns. This setting is especially aligned with our failure-driven curation objective: failed trajectories reveal missing coordination patterns, redundant turns, and brittle decision-making that can be targeted by expansion, filtering, and modification.

    \subsection{Baseline}\label{app:experiment_baseline}
        \paragraph{MUA-RL}
            MUA-RL is a multi-turn user-interacting reinforcement learning method for agentic tool use~\cite{zhao2025mua}. It introduces LLM-simulated users into the RL loop so that the agent learns to clarify user intent, communicate with dynamic users, and invoke tools under uncertain multi-turn demands. We include MUA-RL as a baseline because it represents a data acquisition strategy that improves agent training by enriching interactive rollouts. In contrast, \ourmethod does not rely on a fixed simulated-user pipeline; instead, it uses held-out failures to evolve the data curation code, deciding when to expand, filter, or modify trajectories.
        \paragraph{EnvScaler}
            EnvScaler scales tool-interactive training environments through programmatic synthesis~\cite{song2026envscaler}. It constructs diverse environment skeletons, generates task scenarios, and builds rule-based validation functions, allowing agents to collect trajectories in executable and stateful sandboxes. This baseline evaluates whether simply increasing the amount and diversity of synthetic environments can provide a sufficient training signal for downstream agent generalization. Compared with EnvScaler, \ourmethod focuses on evolving the curation procedure over a given raw corpus, using dev-set failures to select high-utility data, repair weak supervision, and reduce redundant interaction turns.
        \paragraph{AWM}
            Agent World Model (AWM) synthesizes executable, database-backed tool-use environments for large-scale agentic RL~\cite{wang2026agentworldmodelinfinity}. By providing many code-driven environments with reliable state transitions and accessible rewards, AWM enables agents to learn from broad synthetic interaction experiences. We use AWM as a baseline for world-model-style environment scaling, where the main source of improvement comes from generating more trainable environments. \ourmethod differs in that it treats the data processing code itself as the optimization target, thereby adapting SFT data, RL data, and memory construction according to observed agent failures.
        \paragraph{RODS}
            RODS is a reward-driven online data synthesis framework for multi-turn tool-use agents~\cite{fang2026rodsrewarddrivenonlinedata}. It observes that informative RL gradients concentrate near the agent's evolving capability boundary, where rollout outcomes exhibit high reward variance. Based on this insight, RODS uses progress rewards from RL rollouts to detect boundary seed tasks, synthesizes structurally similar multi-turn variants through skill-aligned resampling, and maintains a dynamic replay buffer that co-evolves with the policy. We include RODS as a baseline because it represents an adaptive data expansion strategy that couples data generation with the RL training loop. In contrast, \ourmethod does not primarily optimize online boundary expansion or replay-buffer management; instead, it uses held-out failures to evolve the data curation code itself, enabling the pipeline to select, repair, and transform data across SFT, RL, and memory construction.
        \paragraph{FunReason-MT}
            FunReason-MT is a data synthesis framework for complex multi-turn function calling and agentic tool use~\cite{xu2025funreasonmttechnicalreportadvanced}. It addresses the difficulty of constructing high-quality multi-turn tool-use trajectories through three components: Environment-API Graph Interactions for collecting dependency-aware execution traces, Advanced Tool-Query Synthesis for generating challenging tool-use queries, and Guided Iterative Chain for refining reasoning traces with iterative feedback. We use FunReason-MT as a baseline because it represents a strong offline synthesis pipeline that improves agent training by generating large-scale, logically coherent function-calling data. Compared with FunReason-MT, \ourmethod does not depend on a fixed offline generator or a predefined trajectory synthesis recipe; instead, it adaptively rewrites the data processing procedure according to observed dev-set failures, allowing the curated data distribution to evolve with the target model's weaknesses.

\section{Evolved Curation Code}\label{app:evolved_curate_code}
    This section summarizes the concrete data curation strategy used by the final evolved result, focusing on the design choices that are not fully captured by conventional static filtering or scale-oriented data selection. 
    Instead of treating all retained trajectories as homogeneous training signals, the final strategy explicitly separates policy behaviors, task-specific knowledge, and low-value interactions, then assigns them to different roles in the agent training pipeline. 
    The central idea is to maximize the density of transferable decision signals.
    Behaviors that can improve the agent policy are kept for SFT or RL.
    Long-tail information that should not be memorized by model parameters is converted into retrievable memory.
    Noisy fragments that may dilute preference learning are removed. 
    This yields a compact but behaviorally targeted training set, where the retained data is selected not only for quality but also for its expected contribution to difficult agentic decision states.
    \paragraph{Fine-grained trajectory decomposition.}
        The final curation strategy does not use the full trajectory as the minimum selection unit. A long agent trajectory may contain useful recovery actions, benchmark-specific knowledge, redundant observations, and misleading tool calls at the same time. Therefore, the curation process decomposes trajectories into smaller decision-centric segments and assigns different segments to different uses. This is more suitable for agent data than trajectory-level filtering because the most valuable supervision often appears only at a few critical turns, such as after an invalid observation, a failed precondition, or a state transition.
    \paragraph{Separation between policy data and memory data.}
        After decomposition, the key distinction is whether a segment teaches a transferable policy behavior or mainly contains task-specific knowledge. Transferable behaviors, such as valid tool-call formatting, state-aware continuation, error recovery, and correct termination, are retained as policy-training data. In contrast, long-tail information such as tool availability, function signatures, state constraints, and task-family-specific evidence is compressed into memory rather than directly used for SFT or RL. This separation prevents the model from overfitting to narrow knowledge while still allowing useful contextual hints to be retrieved when they strongly match the current benchmark, task family, and schema-level anchors.
    \paragraph{Failure-family-oriented training signal construction.}
        The retained policy data is further reshaped according to high-frequency failure families observed in model predictions. Instead of uniformly increasing the data scale, the curation strategy increases the density of examples around difficult agentic behaviors, including state-transition errors, invalid function signatures, recovery after failed observations, multi-turn context forgetting, improper use of memory evidence, premature termination, and object-binding errors in service-style tasks. SFT examples are mainly used to stabilize basic formats and reliable recovery patterns, while GRPO groups are selected to provide clear positive-negative contrasts under similar states. In this way, RL optimization focuses on learning why one next action is better than a nearby alternative, rather than simply imitating historical trajectories.
    \paragraph{Complementary use of synthetic data, real trajectories, and discarded data.}
        Synthetic data is used as a targeted curriculum for recurring failure patterns because it provides short and clear contrastive signals. Real trajectories are kept to preserve natural multi-turn structure, noisy observations, context drift, and non-template recovery behavior. At the same time, discarding data is treated as an active part of the strategy rather than a by-product of filtering. Fragments with vague goals, weak decision signals, ambiguous tool calls, excessive irrelevant context, or potentially harmful behaviors are removed to avoid wasting LoRA capacity and diluting GRPO preferences. As a result, the final checkpoint is trained on a smaller but more concentrated distribution that emphasizes transferable agent behaviors and hard decision states.

\section{Failure Mode Analysis}\label{app:fail_mode_analysis}
    This section summarizes the major failure modes observed during the data-curation evolution process. 
    Overall, the failures gradually shift from shallow tool-use issues, such as format mistakes and simple routing errors, to more fundamental agentic bottlenecks. 
    After ordinary tool calling becomes more reliable, the remaining failures mainly concentrate on three capabilities:
    \paragraph{State Binding Drift.}
        We define \emph{State Binding Drift} as the failure to maintain and update correct bindings between task states, entities, parameters, and executable targets across multi-step interactions. Typical manifestations include using stale identifiers, confusing active objects such as reservations or orders, losing cross-turn parameters, or editing an incorrect code location. These errors arise because trajectory-level supervision often rewards locally plausible actions without explicitly teaching persistent state management or observation-conditioned rebinding. Future work could mitigate this issue by curating state-centric trajectories, adding contrastive examples for valid and invalid bindings, and introducing lightweight belief-state or verifier modules to check whether the selected object or parameter is still supported by the latest observation.
    \paragraph{Adaptive Recovery Failure.}
        We define \emph{Adaptive Recovery Failure} as the inability to transform an execution failure, a missing prerequisite, or an environment blocker into an appropriate repair action. Instead of changing strategy, the model may repeat the same failed tool call, ask for information that is already available in the environment, execute writes before confirmation, or stop before the required user-visible effect is achieved. This failure mode is caused by the scarcity of recovery-oriented supervision: successful trajectories mainly demonstrate clean paths, while sparse final rewards provide limited guidance on how to react to intermediate failures. Future work should therefore curate more failure-recovery trajectories, provide step-level feedback for invalid repetitions and premature termination, and use execution monitors to detect loops, unmet prerequisites, and unsafe write-before-confirmation behavior.
    \paragraph{Grounded Execution Gap.}
        We define \emph{Grounded Execution Gap} as the gap between producing a fluent or syntactically plausible action and producing one that is actually supported by evidence, memory, schema constraints, or executable validation. This includes failing to use memory as evidence, extracting unsupported answers, hallucinating tool calls under weak context, violating schema-specific signatures, or generating patches that cannot be applied. The root cause is that many training examples emphasize output format and task completion more than the verification process that links evidence to action validity. Future work could reduce this gap through evidence-linked supervision, verifier-guided data refinement, and execution-aware filtering that retains examples only when the action is both well-grounded and executable.

\end{document}